\documentclass[runningheads]{llncs}

\pdfoutput=1

\usepackage[T1]{fontenc}

\usepackage{graphicx}

\usepackage[breaklinks=true]{hyperref}
\usepackage{breakcites}
\usepackage{color}

\usepackage{xspace}
\usepackage{pifont}
\usepackage{amsmath}
\usepackage{booktabs}
\usepackage{amsfonts}

\newcommand{\cmark}{\ding{51}}
\newcommand{\xmark}{\ding{55}}

\makeatletter
\DeclareRobustCommand\onedot{\futurelet\@let@token\@onedot}
\def\@onedot{\ifx\@let@token.\else.\null\fi\xspace}
\def\eg{\emph{e.g}\onedot} 
\def\ie{\emph{i.e}\onedot}

\makeatother

\usepackage[dvipsnames]{xcolor}
\newcommand{\method}{SITUATE\xspace}

\begin{document}

\title{SITUATE: Indoor Human Trajectory Prediction through Geometric Features and Self-Supervised Vision Representation}

\titlerunning{\method{}: Indoor Human Trajectory Prediction}

\author{
Luigi Capogrosso\inst{1} \and
Andrea Toaiari\inst{1}   \and
Andrea Avogaro\inst{1}   \and
Uzair Khan\inst{1}       \and\\
Aditya Jivoji\inst{2}    \and
Franco Fummi\inst{1}     \and
Marco Cristani\inst{1}
}

\authorrunning{Capogrosso L. et al.}
\institute{Dept. of Engineering for Innovation Medicine, University of Verona, Verona, Italy
\email{name.surname@univr.it} \and
Visvesvaraya National Institute of Technology, Nagpur, India
\email{adjivoji@students.vnit.ac.in}
}

\maketitle

\begin{abstract}
Patterns of human motion in outdoor and indoor environments are substantially different due to the scope of the environment and the typical intentions of people therein.
While outdoor trajectory forecasting has received significant attention, indoor forecasting is still an underexplored research area.
This paper proposes \method{}, a novel approach to cope with indoor human trajectory prediction by leveraging equivariant and invariant geometric features and a self-supervised vision representation.
The geometric learning modules model the intrinsic symmetries and human movements inherent in indoor spaces.
This concept becomes particularly important because self-loops at various scales and rapid direction changes often characterize indoor trajectories.
On the other hand, the vision representation module is used to acquire spatial-semantic information about the environment to predict users’ future locations more accurately.
We evaluate our method through comprehensive experiments on the two most famous indoor trajectory forecasting datasets, \ie{}, TH\"{O}R and Supermarket, obtaining state-of-the-art performance.
Furthermore, we also achieve competitive results in outdoor scenarios, showing that indoor-oriented forecasting models generalize better than outdoor-oriented ones.
The source code is available at \url{https://github.com/intelligolabs/SITUATE}.
\end{abstract}

\section{Introduction}
\label{cha:intro}

Human trajectory prediction is the task of predicting the likely path that a subject will take to reach its designated endpoint~\cite{rudenko2020human}.
This predictive process finds its applicability and utility in a multitude of domains~\cite{kothari2021human}.
For example, in the context of robotics, it serves as a tool for facilitating the predictions on potential future robot trajectories, useful for intelligent planning considering human responses~\cite{salzmann2020trajectron++}.
In industry, human trajectory prediction becomes critical for optimizing automated systems and ensuring seamless interactions with other occupants and components of a production line~\cite{sampieri2022pose}.

Despite the significant volume of research over the past decade devoted to outdoor trajectory prediction~\cite{alahi2016social,gupta2018social,salzmann2020trajectron++,giuliari2021transformer,gu2022stochastic,xu2022remember,bae2022learning}, there has been a notable scarcity of studies that exploited user trajectory data in indoor settings~\cite{mantini2014human,rudenko2020thor,rossi2021human,wang2021location,skenderi2021dohmo,wang2022indoor,toaiari2023scene}, also considering the crucial role these predictions play nowadays in the development of location-based services within indoor spaces.
This gap in research inspired this work, which investigates a learning framework designed explicitly for indoor trajectory prediction.

\begin{figure*}[!t]
\centering
\includegraphics[width=.95\linewidth]{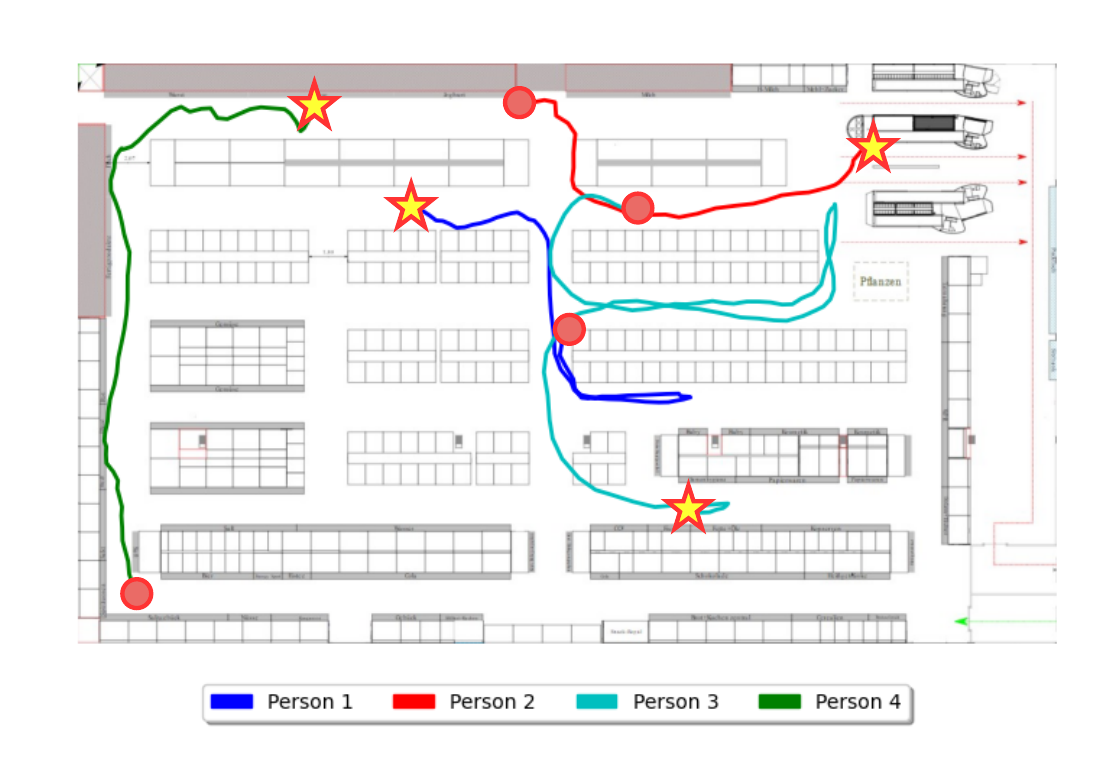}
\caption{Examples of different trajectories from the Supermarket~\cite{gabellini2019large} dataset to show the difficulty of the indoor trajectory prediction task.
In particular, the dataset showcases long trajectories (Person 4), self-loops (Person 1 and Person 3), and confusing movements (Person 2) performed in an environment that strongly affects the people's paths.
Specifically, the red circle represents the starting point of a trajectory, and the yellow star represents its final point.}
\label{fig:indoor_traj_examples}
\end{figure*}

\paragraph{\textbf{Motivations for this paper.}}
In Figure~\ref{fig:indoor_traj_examples}, we can note the distinctive nature of indoor settings, where users can encounter numerous choices and potential pathways.
This factor implies that the dynamic of the motion can be strongly influenced by the environment setup~\cite{mantini2014human,capogrosso2022toward}.
Users can navigate through different interconnected rooms, corridors, doors, and elevators, often having the freedom to deviate from straightforward paths and choose alternative routes.
Indoor spaces also have a higher density of structural elements and potential obstacles, such as furniture, walls, and partitions, as shown in Figure~\ref{fig:indoor_traj_examples}, related to the Supermarket~\cite{gabellini2019large} dataset.
Outdoor environments provide more open spaces, where visibility is less restricted, and the impact of physical barriers is typically reduced~\cite{choi2019looking}.

Consequently, indoor trajectory prediction requires a deeper understanding of the context and semantics of the indoor space, as users may have specific goals, like finding a particular room, reaching a specific point of interest, or accessing various facilities~\cite{guo2017novel}.
This contextual richness adds a layer of complexity to the prediction process since it also makes it necessary to consider the space's physical layout.
Considering the omnipresence of indoor environments in human lives, it is imperative to address trajectory forecasting in these situations.
Indeed, recent studies show that humans spend most of their time in indoor environments such as homes, supermarkets, airports, conference facilities, and train stations~\cite{ashrae2010guideline}.
These considerations form the basis of the research conducted in this work.

\paragraph{\textbf{Innovations in this paper.}}
While outdoor trajectory forecasting has received significant attention, indoor forecasting is still an underexplored research area.
As a result, we present \method{}, the first model designed specifically to cope with indoor trajectory forecasting by leveraging equivariant and invariant geometric feature learning and a self-supervised vision representation.
Taking inspiration from~\cite{xu2023eqmotion}, the equivariant and invariant geometric learning modules were employed to accurately represent intrinsic movements, like self-loops at various scales and hierarchies inherent in indoor spaces.
On the other hand, the self-supervised vision representation module enabled us to acquire spatial-semantic information about the environment, using the scene or space layout images when available, to predict users’ future locations meaningfully and accurately.

In summary, the main contributions of this paper are:
\begin{itemize}
\item We present \method{}, a novel approach for indoor human trajectory forecasting based on equivariant and invariant geometric feature learning modules and a self-supervised vision representation;
\item The equivariant and invariant modules are used to cope with the problem related to the more complicated movements inherent in indoor spaces;
\item The vision representation module is used to acquire spatial-semantic information about the environment to predict users’ future locations more accurately;
\item \method{} also achieves competitive results in outdoor scenarios, showing that indoor forecasting models generalize better than outdoor-oriented ones.
\end{itemize}

\section{Related Work}
\label{cha:related}

\paragraph{\textbf{Indoor human trajectory prediction.}} \label{subsec:indoor_traj_pred}
Predicting the evolution of a pedestrian trajectory in the future is a long-standing task whose interest is constantly renewed by the emergence of new scenarios that can benefit from it, \eg{}, autonomous driving~\cite{rudenko2020human}.
When proposing a methodology to tackle trajectory forecasting, one should take care of several aspects, from the environment's geometry~\cite{mantini2014human} to the presence of obstacles~\cite{haddad2019situation} and the possible interactions between multiple agents~\cite{aydemir2023adapt}.
Some traditional methods to approach this task involved force models~\cite{helbing1995social}, Markov models~\cite{kitani2012activity}, and RNNs~\cite{zhang2019sr}.
Notably, considering common sense rules and conventions that humans observe in social spaces helps to manage simultaneous predictions in crowded scenes~\cite{alahi2016social}.

Multiple deep learning-based models have been applied successfully to forecast pedestrian trajectories, such as GNNs~\cite{yu2020spatio}, Trasformers~\cite{giuliari2021transformer} and Conditional Variational AutoEncoders (CVAEs)~\cite{mangalam2020not}.
More recently, diffusion models have also been applied to solve this problem~\cite{gu2022stochastic}.
However, most proposed methods are tested only on datasets representing outdoor scenarios.
This is due to a lack of comprehensive indoor datasets and the fact that indoor trajectories can be considered more ``difficult'' or non-linear~\cite{rossi2021human}.
When traversing indoors, our immediate movement decision is influenced by the objects in our path and the surrounding walls~\cite{mantini2014human}.
In indoor settings, people navigate in loosely constrained but cluttered spaces with multiple goal points that can be reached in many ways~\cite{rudenko2020thor}.
Moreover, people in indoor scenarios tend to focus on their surroundings, fixating on the most interesting parts of the scene, alternating movement and stationary phases~\cite{toaiari2023scene}.
At the same time, outside, the movement area can be much larger, and the subjects can move further apart.

Some works have been proposed to address the specific problem of indoor trajectory forecasting~\cite{mantini2014human,rudenko2020thor,rossi2021human,skenderi2021dohmo,wang2022indoor}, highlighting the differences between indoor and outdoor trajectory forecasting.
In~\cite{rossi2021human}, the authors address the problem of generalizability, proposing a novel indoor dataset and new metrics to normalize common biases. They tackle the problem of aleatoric multimodality with the GAN-Tri model, which uses a heuristic to produce samples corresponding to different behaviors.
\cite{mantini2014human} examine trajectories, modeled as a Markov chain, within 3D environments, introducing the concept of an occupancy map to represent the relative accessibility of each point on the map with respect to its geometry.
The study emphasizes the importance of proximity from each point to the destination and the occupancy frequency in constructing a probability transition matrix for trajectory prediction.
Unlike them, our approach considers indoor spaces' detailed scene layouts and non-trivial human movements.

\paragraph{\textbf{Equivariant and invariant graph neural networks.}} \label{subsec:eq_inv_models}
Inspired by the research on rotation-equivariant convolutional neural networks within the 2D image domain~\cite{cohen2016group}, the advent of Graph Neural Network (GNN) architectures opened doors to investigating symmetries beyond rotations~\cite{yang2020factorizable}.
For example, in~\cite{sanchez2020learning}, the authors proposed partial equivariance by focusing on translation equivariance.
Meanwhile,~\cite{fuchs2020se} constructed filters using spherical harmonics, enabling equivariance to rotations and translations and facilitating transformations between higher-order representations.

In~\cite{satorras2021n}, a new model for learning equivariant graph neural networks, dubbed EGNNs, is proposed.
Differently from the previous works, this formulation maintains the flexibility of GNNs while remaining E(n) equivariant (translation, rotation, and reflection equivariant) without the need to compute expensive higher-order operations.
\cite{huang2021equivariant} further extended this concept by incorporating geometrical constraints implicitly encoded in the forward kinematics when tackling molecular dynamics prediction and human motion capture.
However, a significant limitation of current methods is their focus solely on state prediction, preventing models from effectively using sequence information.

Recently, EqMotion~\cite{xu2023eqmotion} extended on these ideas to propose an equivariant motion prediction parametric network with an invariant interaction reasoning module, able to tackle distinct problems such as particle and molecule dynamics, human pose forecasting, and outdoor pedestrian trajectory prediction.
Interaction invariance is fundamental in ensuring the agents' interactions remain constant under input transformation.

In our research, we adapt some of the concepts presented in~\cite{xu2023eqmotion} to propose an equivariant model for the human trajectory prediction task, which, in combination with a module to extract semantic information about the scenes, unlock more precise forecasting capabilities in indoor settings.

\paragraph{\textbf{Self-supervised vision representation.}} \label{subsec:scene_understanding}
One way to get image representations without heavily relying on annotated data is to perform Self-Supervised Learning (SSL).
In a nutshell, SSL learns deep feature representations invariant to sensible transformations of the input data.
Then, the learned representations could be used in supervised downstream tasks.

The self-supervised vision representation state-of-the-art rapidly evolved, with Transformer-based architectures emerging as leading models.
The Vision Transformer (ViT)~\cite{dosovitskiy2020image}, and its variants like DeiT~\cite{touvron2021training}, have demonstrated impressive performance in learning powerful visual representations from unlabeled data.
Specifically, these models leverage self-attention mechanisms to capture global context and long-range dependencies within images, enabling them to encode rich semantic information efficiently.

In this paper, to extract semantic information from scenes represented in a 2D map, we use the pre-trained BEiT~\cite{bao2021beit}, the state-of-the-art self-supervised vision representation model.
This offers a powerful framework for learning visual representations without explicit supervision, effectively capturing high-level semantics and intricate features inherent in visual data.

\section{Method} 
\label{cha:method}

\paragraph{\textbf{Mathematical background.}} \label{subs:background}
Given a set of transformations $T_x : X \rightarrow X$, a function $F : X \rightarrow Y$ is called Equivariant if exists a transformation $T_y : Y \rightarrow Y$ equivalent to $T_x$, on the Euclidean space such that:
\begin{equation}
F(T_x(X))=T_y(F(X))\;.
\end{equation}
Moreover, we also want the model to have the invariance property.
Given the same set of transformations, a function $F : X \rightarrow Y$ is called Invariant on the Euclidean space if it exists a transformation $T_y : Y \rightarrow Y$ such that:
\begin{equation}
F(X)=F(T_x(X))\;.
\end{equation}

Specifically, this work addresses the problem of multi-person trajectory forecasting by considering the input trajectories as a graph.
As proven by~\cite{satorras2021n}, during the message passing of a GNN, the property of equivariance can be ensured by enriching the features of the neighbor nodes with the $L2$ distance between nodes.
Let $G=\{V, E\}$ be an input graph representing the input trajectory with nodes $v_i \in V$ and edges $e_{ij} \in E$.
For every node $v_i$, a feature vector $h \in \mathbf{R}^h$ and an absolute position $x_i \in \mathbf{R}^3$ are given.
To preserve equivariance among different layers of the model, we update the position as follows:
\begin{align}
m_{ij} &=\phi_{e}\left( h_{i}^{l}, h_{j}^{l}, \left\|x_{i}^{l}-x_{j}^{l}\right\|^{2}\right) \label{eq:modules_input}\;, \\
x_{i}^{l+1} &=x_{i}^{l}+ C\sum_{j \neq i}\left(x_{i}^{l}-x_{j}^{l}\right) \phi_{x}\left(m_{ij}\right)\;,
\label{eq:equivariance}
\end{align}
where $C$ is equal to $1/(M-1)$ with $M$ number of nodes,
$\phi_{e}$ and $\phi_{x}$ are learnable Multi-layer Perceptrons (MLPs), defined as $\phi_e(\cdot)=W_e \cdot + B_e$, $l$ indicates the layer and
$m_{ij}$ represents the information passed between two nodes during the message passing.

As reported in~\cite{satorras2021n}, $\phi_{x}$ has to be a scoring function $\phi_{x} : X \rightarrow S$, with $S \in \mathbf{R}^1$.
With this procedure, the update of $V$ is consistent, allowing the model to learn without being affected by SO(2) transformations, with SO(2) being the group of all rotations in the plane around the origin that preserve the Euclidean norm,  mathematically described by $2\times2$ matrices.
Furthermore, the features learned across layers must be consistent and invariant to graph transformations.
To do so, the following procedure governs the final message-passing operations and the update of the features carried out by the $i-th$ layer:
\begin{align}
m_{i} &=\sum_{j \neq i} m_{ij}\;, \\ 
h_{i}^{l+1} &=\phi_{h}\left(h_{i}^l, m_{i}\right)\;,
\label{eq:invariance}
\end{align}
with $\phi_{h}$, an MLP also designed as $\phi_h(X)=W_h X + B_h$, responsible for the invariant feature learning.
Mixing these two components allows us to build an Equivariant and Invariant GNN using Euclidean SO(2) transformations.

\paragraph{\textbf{Motion prediction.}} \label{subs:formulation}
Here, we introduce the general problem formulation of motion prediction.
We have a multi-agent system with $m$ agents.
Each agent is represented as $A_i$, where $i = 1, 2, \ldots, m$. The goal is to predict the future motions of these agents based on their historical observations.
For each agent $A_i$, we can denote the historical observations as $X_i$.
These observations typically include positions and can be represented as $X_i = \{x^{i}_0, x^{i}_1, \ldots ,x^{i}_t\}$, where $x^i_t$ represents the position of agent $A_i$ at time step $t$.
We also add the velocity $S_i = \{s^{i}_{t+1}, s^{i}_{t+2}, \ldots ,s^{i}_{t+f}\}$ as input information of the model.
The velocity of an agent is a natural invariant feature because it is not affected by any translation or SO(2) transformation.
We use the velocity to compute the initial feature vector of a specific agent $A_i$.
More details in Section~\ref{subsec:feature_init}.
Specifically, for each agent $A_i$ we aim to predict its future $f$ positions $Y_i = \{y^{i}_{t+1}, y^{i}_{t+2}, \ldots ,y^{i}_{t+f}\}$.

\begin{figure*}[!t]
\centering
\includegraphics[width=\linewidth]{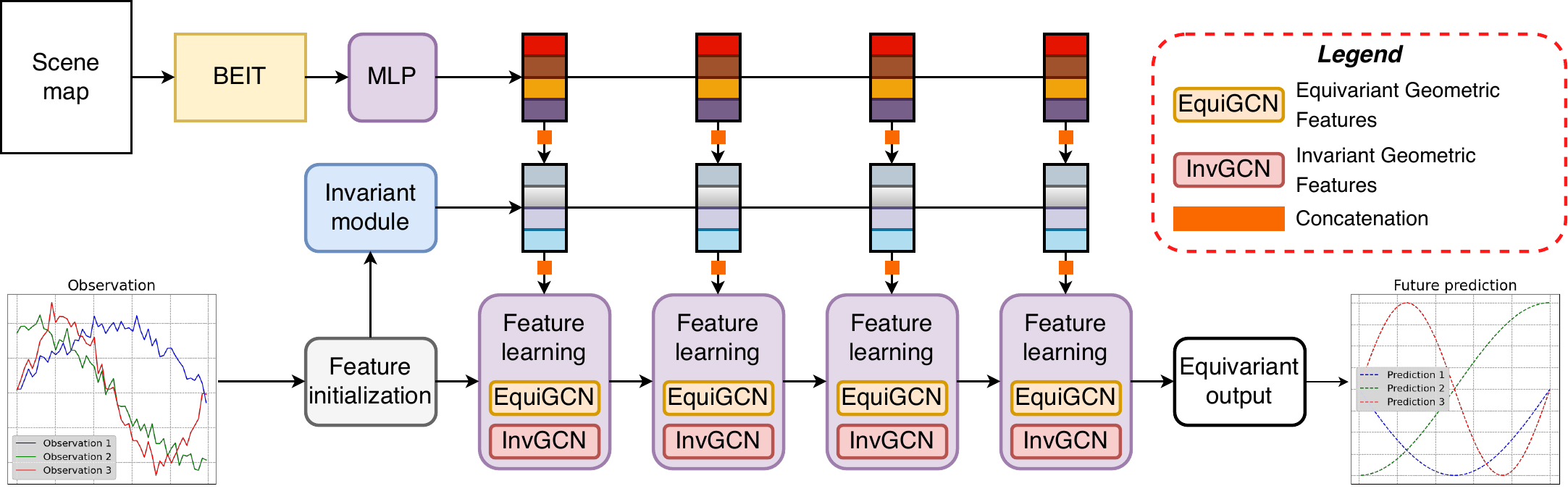}
\caption{In \method{}, we first produce a feature vector regarding the scene using the self-supervised vision representation module.
Then, a feature initialization layer is used to initialize geometric and pattern features.
We then successively update the geometric and pattern features by the equivariant geometric feature learning and invariant pattern feature learning layers, obtaining expressive feature representation.
We further use an invariant reasoning module to infer an interaction graph used in equivariant geometric feature learning.
Finally, we use an equivariant output layer to obtain the final prediction.}
\label{fig:architecture}
\end{figure*}

\subsection{The \method{} Prediction Network} \label{subs:eq_in}
In this section, we present \method{}, our motion prediction network that explicitly uses equivariant and invariant geometric features and a self-supervised scene representation module to tackle the indoor trajectory prediction problem.
The model architecture is shown in Figure~\ref{fig:architecture}.

The first module we present is in charge of producing the scene-representation encoding.
As anticipated, the subjects' motion characteristics differ greatly from those of the outdoor case when considering indoor trajectory forecasting.
Indeed, the motion is strongly characterized and limited by the objects and obstacles in the scene.
Knowing the available space that limits the viable paths in the scene can, for every $X_i$, strongly reduce the cardinality of all the possible outcomes of the model.
Starting from the assumption that all the scene objects and structure are available in the form of a scene layout or a camera image, BEiT~\cite{bao2021beit} is first used to output visual tokens $T_s$, the so-called scene-representation encodings.
These tokens $T_s$ are fed into a learnable MLP defined as $\phi_t : T_s \rightarrow T_e$ and then concatenated into the input.

The input concatenated with $T_e$ is then fed into two modules: Equivariant block ($EquiGCN$) and Invariant block ($InvGCN$).
Following~\cite{xu2023eqmotion}, these two blocks are both based on the implementation of the message passing described in Equation~\ref{eq:modules_input}, modified to accept also $T_e$:
\begin{align}
m_{ij} &=\phi_{e}\left(h_{i}^{l}, h_{j}^{l}, \left\|x_{i}^{l}-x_{j}^{l}\right\|^{2}, T_e, a_{i j}\right)\;,
\label{eq:modules_input_token}
\end{align}
where $a_{i j}$ is the edge attribute (or weight), which can be derived from the adjacency matrix.

Specifically, the $EquiGCN$ block is responsible for the update of the node's coordinates $x$, and it represents the implementation of the update function described in Equation~\ref{eq:equivariance}.
On the other hand, $InvGCN$ is the implementation of the update function of the node's features $h$ in Equation~\ref{eq:invariance}.

The possible pathways are learned by the module $\phi_t$, starting from the token produced by the pre-trained BeIT model.
The outputs of $EquiGCN$ and $InvGCN$ are computed as reported respectively in Equation~\ref{eq:equivariance} and Equation~\ref{eq:invariance}, updating $h_{i}^{l+1}$ and $x_{i}^{l+1}$.

To understand the contribution of these two modules in less formal and more practical terms, imagine a navigation system.
It can get from point A to point B but might struggle with tricky situations.
In \method{}, EquiGCN injects a sense of direction like a compass you wear on your hat.
Specifically, it ensures the network understands the environment's layout, regardless of where it starts ``looking''.
InvGCN, on the other hand, acts like a map you hold – it helps the network account for different starting points and body orientations, making the predictions more robust.

\subsection{Feature Initialization} \label{subsec:feature_init}
The input given to our model is a set of trajectories of different agents.
The first step is to define a node for every position $x_t$ for every agent $A_i$.
Every node $x_t$ is connected with $x_{t-1}$ and $x_{t+1}$ if they are related to the same agent $A_i$.
Since only trajectories (and thus positions $x_t$) are given as starting data, it is necessary to define for each trajectory a vector of initial features $h^{0}_i$ to be used as input together with the positions $x^{0}_i$.

As stated in~\cite{huang2021equivariant}, having an invariant feature vector $h^0_i$ is necessary to guarantee equivariance.
Given that as input data we only have position $X_i$, we followed the  procedure in~\cite{xu2023eqmotion} to use velocities in order to create $h^0_i$ as follows:
\begin{gather}
\hat{x}_i = \phi_{X}( X_i + \overline{\mathbb{H}} ) + \overline{\mathbb{H}}\;, \\
\rho^{t}_i = \| v^t_i\|_2\;, \\
\theta_i^{t} = \text{angle}(v_i^{t},v_i^{t-1})\;, \\
h_i^{0} = \phi_{h_0}(\rho_i,\theta_i)\;,
\label{eq:feat_init}
\end{gather}
where $h_i^{0}$ is the initial features vector of the $i-th$ agent.
$v^t_i$ represent the velocity of the agent and is defined as $\bigtriangleup \hat{x}_i^{t}$, where
$\bigtriangleup $ is the finite difference operator, $\overline{\mathbb{H}}$ is the centroid of the observed trajectories of all agents in the scene.
$\phi_{X}$ and $\phi_{h_0}$ are two fully connected layers responsible for encoding and producing the initial graph and the initial features of the trajectory.

To compute $h_i^{0}$, two different types of velocities are needed (thus, information invariant to rotation and translation): $\bigtriangleup x_i^{t}$ effectively represents the Euclidean velocity of the agent and $\theta_i^{t}$ represents the angular velocity on a certain time step $t$.
Note that both $\phi_{x_0}$ and $\phi_{h_0}$, and in general all the operations described, are linear transformations: this is necessary to preserve both equivariance and invariance properties of the remaining part of the model.

\section{Experiments}
\label{cha:experiments}

Our experimental evaluation is tailored toward two objectives.
Firstly, in Section~\ref{subsec:indoor_res}, we show the superiority of \method{} in the two most well-known indoor datasets, defining the new state-of-the-art in indoor scenarios.
Secondly, in Section~\ref{subsec:outdoor_res}, we prove that \method{} can also achieve comparable results with respect to other competitors on outdoor datasets.
Finally, in Section~\ref{subsec:ablation}, we report some ablation studies.

\subsection{Evaluation Setup}

\paragraph{\textbf{Datasets.}}
We evaluate \method{} on the state-of-the-art indoor datasets and the most well-known outdoor human trajectory prediction dataset.

\noindent{\textbf{TH\"{O}R.}} The TH\"{O}R dataset~\cite{rudenko2020thor} includes human motion trajectory and gaze data collected in an indoor environment with accurate ground truth for the participants' position.
It comprises 395K frames at 100 Hz, 2531K people detections, and over 600 individual and group trajectories between multiple resting points.
The map was taken from the dataset's official website.

\noindent{\textbf{Supermarket.}} The Supermarket dataset~\cite{gabellini2019large} comprises 4 different scenario: German1, German2, German3, and German4, \ie{}, four different supermarket.
The dataset collection involved attaching devices on shopping carts/baskets and recording their movements during customer usage.
Each subset includes a file with a map of the supermarket.

\noindent{\textbf{ETH-UCY.}} The ETH~\cite{pellegrini2010improving} and UCY~\cite{lerner2007crowds} dataset group consists of five different scenes: ETH \& HOTEL (from ETH) and UNIV, ZARA1, \& ZARA2 (from UCY).
The scenes are captured in unconstrained outdoor environments with few objects blocking the pedestrian paths.
In this case, images of the scene were used.

\paragraph{\textbf{Evaluation metrics.}}
We use standard metrics for the trajectory prediction task, \ie{}, minimum Average Displacement Error (ADE), and minimum Final Displacement Error (FDE).
In particular, ADE measures the average $L2$ difference between the prediction at all time steps and the ground truth.
On the other hand, FDE measures the difference between the predicted endpoint and the ground truth.

\paragraph{\textbf{Prediction mode.}}
Following the evaluation protocol of~\cite{xu2023eqmotion}, \method{} is employed in two prediction modes: deterministic and multi-prediction.
Deterministic means the model only outputs a single prediction for each input motion observation, while multi-prediction means the model has 20 predictions for each input motion observation.
Under multi-prediction, ADE and FDE will be calculated using the best-predicted trajectory.
To adapt to multi-prediction, we modify \method{} to repeat the last feature updating layer and the output layer 20 times in parallel to have a multi-head prediction.

\paragraph{\textbf{Implementation details.}}
As a backbone for our model, we used the structure of~\cite{xu2023eqmotion}.
The model architecture has four layers of geometric feature learning.
We use the SiLU activation function and dropout with a 0.5 probability to regularise within all MLPs.
The visual embeddings of the image, \ie{}, the floor plans (look at Figure~\ref{fig:indoor_traj_examples}) for context information are derived from the last layer of the BEiT model.
The model is provided with past trajectory information spanning eight discrete time steps, and the model's task is to predict 12 steps into the future.
In addition to the dropout mentioned above, we apply the Discrete Cosine Transform (DCT) to the input data as a regularisation technique.
Specifically, by representing the data in the frequency domain, it becomes easier to distinguish between signal and noise components, resulting in a cleaner signal.
The impact of these regularization approaches is discussed in Section~\ref{subsec:ablation}.
We train our models with a batch size of 64 for 60 epochs, using AdamW~\cite{loshchilov2017decoupled} as an optimizer within the PyTorch Lightning framework on an NVIDIA RTX 3090.

\subsection{Indoor Human Trajectory Prediction Results} \label{subsec:indoor_res}
\begin{table}[t!]
\centering
\caption{Deterministic prediction performance (ADE ($m$)/FDE ($m$)) on the TH\"{O}R and the Supermarket datasets. 
The \textbf{bold}/\underline{underlined} font denotes the \textbf{best}/\underline{second-best} result.}

\begin{tabular}{l|c|c||c}
\toprule
& \multicolumn{3}{c}{Performance (ADE ($m$) $\downarrow{}$/ FDE ($m$) $\downarrow{}$)} \\
\midrule
\textbf{Deterministic Evaluation} & \textbf{TH\"{O}R} & \textbf{Supermarket} & \textbf{Average} \\
\midrule
TransF~\cite{giuliari2021transformer}   & 2.62/4.81 & 2.56/2.90 & 2.59/3.85 \\
MemoNet~\cite{xu2022remember}           & 0.78/5.05 & 1.79/2.94 & 1.28/3.99 \\
EqMotion~\cite{xu2023eqmotion}          & \underline{0.56}/\underline{0.94} & \underline{1.71}/\underline{2.65} & \underline{1.13}/\underline{1.79} \\
\method{} (ours)                        & \textbf{0.45}/\textbf{0.93} & \textbf{1.21}/\textbf{1.84} & \textbf{0.83}/\textbf{1.38} \\
\bottomrule
\end{tabular}

\label{tab:indoor_deterministic}
\end{table}
\begin{table}[t!]
\centering
\caption{Multi-prediction performance (ADE ($m$)/FDE ($m$)) on the TH\"{O}R and the Supermarket datasets. 
The \textbf{bold}/\underline{underlined} font denotes the \textbf{best}/\underline{second-best} result.}

\begin{tabular}{l|c|c||c}
\toprule
& \multicolumn{3}{c}{Performance (ADE ($m$) $\downarrow{}$/ FDE ($m$) $\downarrow{}$)} \\
\midrule
\textbf{Multi-prediction Evaluation} & \textbf{TH\"{O}R} & \textbf{Supermarket} & \textbf{Average} \\
\midrule
PECNet~\cite{mangalam2020not}   & --     & 1.57/3.45 & -- \\
GP-Graph~\cite{bae2022learning} & 2.80/3.92 & 3.19/4.57 & 2.99/4.24 \\
EqMotion~\cite{xu2023eqmotion}  & \underline{1.32}/\underline{1.03} & \underline{1.29}/\underline{1.77} & \underline{2.61}/\underline{1.40} \\
\method{} (ours)                & \textbf{0.50}/\textbf{0.86} & \textbf{0.53}/\textbf{0.65} & \textbf{0.51}/\textbf{0.75} \\
\bottomrule
\end{tabular}

\label{tab:indoor_multiprediction}
\end{table}

We conducted comparative experiments to assess the soundness of our approach against existing trajectory prediction methods.
The methods include deterministic evaluation models (TransF~\cite{giuliari2021transformer}, MemoNet~\cite{xu2022remember}), as well as multi-prediction evaluation models (PECNet~\cite{mangalam2020not}, GP-Graph~\cite{bae2022learning}).
Our evaluation for both prediction modes also encompasses EqMotion~\cite{xu2023eqmotion}, the state-of-the-art method with invariant end equivariant interaction reasoning.
Table~\ref{tab:indoor_deterministic} and Table~\ref{tab:indoor_multiprediction} show the results.

The results show that the proposed \method{} consistently outperforms all baseline methods in all cases.
On the TH\"{O}R dataset, \method{} achieves an ADE of 0.45 and an FDE of 0.93, showcasing its superiority over other models.
Notably, compared to EqMotion, the second-best model, \method{} exhibits a substantial 22\% reduction in ADE and a 1\% reduction in FDE.

Similarly, on the Supermarket dataset, \method{} continues demonstrating its effectiveness with an ADE of 1.21 and an FDE of 1.84.
Compared to EqMotion, again the closest competitor, \method{} achieves a 29\% reduction in ADE and a 31\% in FDE, reinforcing its dominance.

The consistently good performance of \method{} across both datasets underscores its robustness and efficacy in trajectory prediction tasks.
The results further suggest that \method{} is accurate and that using scene information when tackling indoor prediction scenarios offers a key advantage compared to the available approaches.

\subsection{Outdoor Human Trajectory Prediction Results} \label{subsec:outdoor_res}
\begin{table}[t!]
\centering
\begin{small}
\caption{Deterministic prediction performance (ADE ($m$)/FDE ($m$)) on the ETH-UCY dataset. 
The \textbf{bold}/\underline{underlined} font denotes the best/second-best result.}

\begin{tabular}{l|c|c|c|c|c||c}
\toprule
& \multicolumn{6}{c}{Performance (ADE ($m$) $\downarrow{}$/ FDE ($m$) $\downarrow{}$)} \\
\midrule
\textbf{Deterministic} & \textbf{ETH} & \textbf{HOTEL} & \textbf{UNIV} & \textbf{ZARA1} & \textbf{ZARA2} & \textbf{Average} \\
\midrule
S-LSTM~\cite{alahi2016social}          & 1.09/2.35 & 0.79/1.76 & 0.67/1.40 & 0.47/1.00 & 0.56/1.17 & 0.72/1.54 \\
SGAN-ind~\cite{gupta2018social}        & 1.13/2.21 & 1.01/2.18 & 0.60/1.28 & 0.42/0.91 & 0.52/1.11 & 0.74/1.54 \\
Traj++~\cite{salzmann2020trajectron++} & 1.02/2.00 & \underline{0.33}/0.62 & \underline{0.53}/\underline{1.19} & 0.44/0.99 & \underline{0.32}/0.73 & \underline{0.53}/\underline{1.11} \\
TransF~\cite{giuliari2021transformer}  & 1.03/2.10 & 0.36/0.71 & \underline{0.53}/1.32 & 0.44/1.00 & 0.34/0.76 & 0.54/1.17 \\
MemoNet~\cite{xu2022remember}          & 1.00/2.08 & 0.35/0.67 & 0.55/\underline{1.19} & 0.46/1.00 & 0.37/0.82 & 0.55/1.15 \\
EqMotion~\cite{xu2023eqmotion}         & \underline{0.96}/\underline{1.92} & \textbf{0.30}/\underline{0.58} & \textbf{0.50}/\textbf{1.10} & \textbf{0.39}/\textbf{0.86} & \textbf{0.30}/\textbf{0.68} & \textbf{0.49}/\textbf{1.03} \\
\method{} (ours)                       & \textbf{0.94}/\textbf{1.90} & \textbf{0.30}/\textbf{0.57} & \textbf{0.50}/\textbf{1.10} & \underline{0.41}/\underline{0.89} & \underline{0.32}/\underline{0.70} & \textbf{0.49}/\textbf{1.03} \\
\bottomrule
\end{tabular}

\label{tab:outdoor_deterministic}
\end{small}
\end{table}

\begin{table}[t!]
\centering
\begin{small}
\caption{Multi-prediction performance (ADE ($m$)/FDE ($m$)) on the ETH-UCY dataset. 
The \textbf{bold}/\underline{underlined} font denotes the best/second-best result.}

\begin{tabular}{l|c|c|c|c|c||c}
\toprule
& \multicolumn{6}{c}{Performance (ADE ($m$) $\downarrow{}$/ FDE ($m$) $\downarrow{}$)} \\
\midrule
\textbf{Multi-prediction} & \textbf{ETH} & \textbf{HOTEL} & \textbf{UNIV} & \textbf{ZARA1} & \textbf{ZARA2} & \textbf{Average} \\
\midrule
SGAN~\cite{gupta2018social}            & 0.87/1.62 & 0.67/1.37 & 0.76/0.52 & 0.35/0.68 & 0.42/0.84 & 0.61/1.21 \\
STGAT~\cite{huang2019stgat}            & 0.65/1.12 & 0.35/0.66 & 0.34/0.69 & 0.29/0.60 & 0.52/1.10 & 0.43/0.83 \\
STAR~\cite{yu2020spatio}               & 0.36/0.65 & 0.17/0.36 & 0.31/0.62 & 0.29/0.52 & 0.22/0.46 & 0.26/0.53 \\
NMMP~\cite{hu2020collaborative}        & 0.61/1.08 & 0.33/0.63 & 0.52/1.11 & 0.32/0.66 & 0.43/0.85 & 0.41/0.82 \\
Traj++~\cite{salzmann2020trajectron++} & 0.61/1.02 & 0.19/0.28 & 0.30/0.54 & 0.24/0.42 & 0.18/0.31 & 0.30/0.51 \\
PECNet~\cite{mangalam2020not}          & 0.54/0.87 & 0.18/0.24 & 0.35/0.60 & 0.22/0.39 & 0.17/0.30 & 0.29/0.48 \\
Agentformer~\cite{yuan2021agentformer} & 0.45/0.75 & 0.14/0.22 & 0.25/0.45 & \underline{0.18}/\textbf{0.30} & \underline{0.14}/\underline{0.24} & 0.23/0.39 \\
GroupNet~\cite{xu2022groupnet}         & 0.46/0.73 & 0.15/0.25 & 0.26/0.49 & 0.21/0.39 & 0.17/0.33 & 0.25/0.44 \\
MID~\cite{gu2022stochastic}            & \textbf{0.39}/0.66 & 0.13/0.22 & \textbf{0.22}/0.45 & \textbf{0.17}/\textbf{0.30} & \textbf{0.13}/0.27 & \textbf{0.21}/0.38 \\
GP-Graph~\cite{bae2022learning}        & 0.43/\underline{0.63} & 0.18/0.30 & 0.24/\textbf{0.42} & \textbf{0.17}/\underline{0.31} & 0.15/0.29 & 0.23/0.39 \\
EqMotion~\cite{xu2023eqmotion}         & \underline{0.40}/\textbf{0.61} & \textbf{0.12}/\textbf{0.18} & \underline{0.23}/\underline{0.43} & \underline{0.18}/0.32 & \textbf{0.13}/\textbf{0.23} & \textbf{0.21}/\textbf{0.35} \\
\method{} (ours)                       & 0.41/0.64 & \underline{0.13}/\underline{0.20} & \underline{0.23}/\underline{0.43} & 0.22/0.35 & \underline{0.14}/0.26 & \underline{0.22}/\underline{0.37} \\
\bottomrule
\end{tabular}

\label{tab:outdoor_multiprediction}
\end{small}
\end{table}

We also evaluate the performance of \method{} with the deterministic and multi-prediction modalities with outdoor scenarios.
Here, we show that \method{} achieves competitive results, showing that indoor-oriented forecasting models tend to generalize better than outdoor-oriented ones.
Table~\ref{tab:outdoor_deterministic} and Table~\ref{tab:outdoor_multiprediction} present the quantitative results.

Specifically, when considering the deterministic prediction case, \method{} demonstrates a performance improvement by obtaining state-of-the-art results in both ADE and FDE across the ETH (0.94/1.90), HOTEL(0.30/0.57), and UNIV (0.50/1.10) scenes.
It places second in the ZARA1 and ZARA2 scenes while performing on par with EqMotion~\cite{xu2023eqmotion} when considering average performance.
In the context of multi-prediction modality, \method{} secures the second rank in terms of ADE and FDE across nearly 75\% of the scenes within the ETH-UCY dataset while maintaining an overall second place in average performance.

Designed for indoor scenarios and their peculiar conformations, \method{} remarkably demonstrates robust capabilities, even when tested on outdoor datasets.
In contrast, this is not always true for architectures tailored for outdoor instances, as we can observe in Table~\ref{tab:indoor_deterministic} and Table~\ref{tab:indoor_multiprediction}, that often struggle when confronted with scenes that differ from those for which they were designed.

\subsection{Ablation Study}\label{subsec:ablation}
\begin{table}[t!]
\centering
\begin{small}
\caption{Ablation results (ADE ($m$)/FDE ($m$)) of \method{}.
We assess the contribution of the scene representation module and regularization methods in the deterministic prediction case.}

\begin{tabular}{cc|c|c||c}
\toprule
& & \multicolumn{3}{c}{Performance (ADE ($m$) $\downarrow{}$/ FDE ($m$) $\downarrow{}$)} \\
\midrule
\begin{tabular}[c]{@{}c@{}}\textbf{Scene}\\ \textbf{Representation}\end{tabular}  & \textbf{Regularization} & \textbf{THOR} & \textbf{Supermarket} & \textbf{Average} \\
\midrule
\xmark & \xmark & 0.50/1.02 & 1.92/1.55 & 1.21/1.29 \\
\xmark & \cmark & 0.56/0.74 & 1.79/2.94 & 1.18/1.84 \\
\cmark & \xmark & 0.57/0.96 & 1.29/1.89 & 0.93/1.43 \\
\cmark & \cmark & \textbf{0.45/0.93} & \textbf{1.21/1.84} & \textbf{0.83/1.38} \\
\bottomrule
\end{tabular}

\label{tab:ablation}
\end{small}
\end{table}

We quantitatively evaluate the impact of the scene representation module and regularization methods by considering the deterministic indoor prediction scenario.
Results are summarized in Table~\ref{tab:ablation}.
It is observed that both contributions play a crucial role in enhancing the overall performance of \method{}.
In particular, the scene representation module effectively encodes semantic information from the visual scene maps, facilitating an accurate understanding of the environment. 
On the other hand, the regularization methods ensure robustness and generalization of the model by mitigating overfitting and improving its ability to generalize to unseen data.
Therefore, we assert that both contributions are indispensable for achieving the desired outcomes and validating the efficacy of our approach.

\section{Conclusion}
\label{cha:conclusions}

This paper presents \method{}, a graph neural network-based model designed specifically to cope with indoor human trajectory prediction.
\method{}, using geometric features and self-supervised vision representations, models the intricate human movements inherent in indoor spaces and accurately predicts users’ future locations.
The scene vision representation module provides insights about the environment, particularly helping in those indoor scenes that are more constrained and full of obstacles.
We evaluate our method on two well-known indoor trajectory prediction datasets, \ie{}, TH\"{O}R and Supermarket, and achieve state-of-the-art prediction performance.
Furthermore, we also achieve competitive results in outdoor scenarios, showing that indoor-oriented forecasting models generalize better than outdoor-oriented ones.

\section*{Acknowledgement}
This study was carried out within the PNRR research activities of the consortium iNEST (Interconnected North-Est Innovation Ecosystem) funded by the European Union Next-GenerationEU (Piano Nazionale di Ripresa e Resilienza (PNRR) – Missione 4 Componente 2, Investimento 1.5 – D.D. 1058  23/06/2022, ECS\_00000043).
This manuscript reflects only the Authors’ views and opinions.
Neither the European Union nor the European Commission can be considered responsible for them.

\bibliographystyle{splncs04}
\bibliography{bibi}

\end{document}